# Person re-identification via efficient inference in fully connected CRF

Wan Jiuqing

Department of Automation

Beijing University of Aeronautics and Astronautics

Beijing, China

wanjiuqing@buaa.edu.cn

Xing Menglin

Department of Automation

Beijing University of Aeronautics and Astronautics

Beijing, China

xingmenglin@buaa.edu.cn

*Abstract*—In this paper, we address the problem of person re-identification problem, i.e., retrieving instances from gallery which are generated by the same person as the given probe image. This is very challenging because the person's appearance usually undergoes significant variations due to changes in illumination, camera angle and view, background clutter, and occlusion over the camera network. In this paper, we assume that the matched gallery images should not only be similar to the probe, but also be similar to each other, under suitable metric. We express this assumption with a fully connected CRF model in which each node corresponds to a gallery and every pair of nodes are connected by an edge. A label variable is associated with each node to indicate whether the corresponding image is from target person. We define unary potential for each node using existing feature calculation and matching techniques, which reflect the similarity between probe and gallery image, and define pairwise potential for each edge in terms of a weighed combination of Gaussian kernels, which encode appearance similarity between pair of gallery images. The specific form of pairwise potential allows us to exploit an efficient inference algorithm to calculate the marginal distribution of each label variable for this dense connected CRF. We show the superiority of our method by applying it to public datasets and comparing with the state of the art.

## I. Introduction

In this paper, we address the problem of person re-identification, which is a standard component of multi-camera surveillance system as it is a way of associate multiple observations of the same individual over time. In realistic, wide-area surveillance scenarios such as airports, metro, and train stations, re-identification systems should be capable of robustly associating a unique identity with hundreds, if not thousands, of individual observations collected from a distributed network of many sensors. There are different settings of person re-identification task [1]. We consider the following one in this work. Suppose picture(s) of a target person is provided, the goal is to find all the corresponding instances belonging to this target in a large video datasets. A typical example of application is to search for and recover the trajectory of a suspect from the video collected by networks of cameras monitoring the region of interest.

An automated re-identification system should be able to (i) generate candidate pedestrian pictures (called gallery of images in this paper) from the video and, (ii) determine which images in the gallery come from the target person by comparing them with the available target picture (called probe image in this paper). We assume that the first sub-problem has been solved using some pedestrian detection algorithms [2] and focus on the second sub-problem, which is still facing several challenges in practice. First, the camera network usually has complicated spatial and temporal topology. Therefore, given a query image of a person, the candidate set could be very large, introducing high levels of uncertainty to person re-identification. Second, there is variability in illumination, camera angles and views, background clutter, and occlusion over the camera network. In particular, the complexity increases in proportion to the scale of the camera network. Finally, the human body is articulated, and a person's appearance can change almost continuously.

Most of the current works about person re-identification follow a common pipeline: first, extracting the imagery features and constructing a descriptor; and second, evaluating the distance (or similarity) between the probe and each element in the gallery according to an appropriate metric. The elements in gallery that similar to the probe are thought to be targets. Motivated by [3], in this paper we take a different angle of view for person re-identification problem. The key assumption underlying our method is that the elements in gallery belonging to the target should not only be similar to the probe, but also be similar to each other, under suitable metric. In other words, comparisons are made between probe and gallery images, and at the same time between each pair of gallery images. We express this assumption with a fully connected CRF model in which each node corresponds to a gallery and each pair of nodes are connected by an edge. A label variable is associated with each node to indicate whether the corresponding image is from target person. We define unary potential for each node using existing feature calculation [7-12] and matching techniques [15-21], which reflect the similarity between probe and gallery image, and define pairwise potential for each edge in terms of a weighed combination of Gaussian kernels, which encode appearance similarity between pair of gallery images. The specific form of pairwise potential allows us to exploit an efficient inference algorithm [4] to calculate the marginal distribution of each label variable for this dense connected CRF. The idea of considering similarity between gallery images is also exploited in [5]. However, the CRF model used in [5] is sparsely connected in which each node is connected with its k-nearest neighbors in feature space. In addition, [5] solve a MAP problem using graph cut algorithm and can only output a hard decision about the label of each gallery image without a confident score of the decision. Another related work is [6], in which the authors take into account the similarity between every pair of images that are consecutive in a "path", and formulate the re-identification problem as a binary integer programming problem with a set of path consistency constraints. However, if the similarity of each pair of images in the gallery were considered, just like that in our fully connected CRF model, the number of path in [6] would become intractable.

## II. REALTED WORKS

Literatures about person re-identification can be generally categorized into two themes.

One of them is focusing on designing suitable feature representation for person re-identification. Ideally, the features extracted should be robust to changes in illumination, viewpoint, background clutter, occlusion and image quality/resolution. Ma [7,8] proposed two kinds of robust images descriptor for these purpose. The first one [7] is based on Biologically Inspired Features extracted through the use of Gabor filters and MAX operator, which are encoded by the covariance descriptor of [9], used to compute the similarity of BIF features at neighboring scales. The second one [8] builds on the combination of recently proposed Fisher Vectors for image classification [10] and a novel and simple seven-dimensional local descriptor adapted to the representation of person images, and use the resultant representation as a person descriptor. Bazzani [11] presented a robust symmetry-based descriptor for modeling the human appearance, which localizes perceptually relevant body parts driven by asymmetry and/or symmetry principles. Specifically, the descriptor imposes higher weights to features located near to the vertical symmetry axis than those that are far from it, permitting higher preference to internal body foreground, rather than peripheral background portions in the image. Layne [12] takes a different view of learning mid-level semantic attribute features reflecting a low dimensional human interpretable description of each person's appearance.

Another theme of works in person re-identification is focusing on optimizing the matching of probe image against images in gallery, to overcome the inter-class confusion and intra-class variation. A number of studies have proposed different ways for estimating the Brightness Transfer Function [13,14], modeling the changes of color distribution of objects transiting from one camera view to another. However, in many cases the transfer functions between camera view pairs are complex and multi-modal. A popular alternative to color transformation learning is distance metric learning. The idea of distance metric learning is to search for the optimal metric under which instances belonging to the same person are more similar, and instances belonging to different people are more different. Existing distance metric learning methods for re-identification include Large Margin Nearest Neighbour [15], Information Theoretic Metric Learning [16], Logistic Discriminant Metric Learning [17], KISSME [18], RankSVM [19], Probabilistic Relative Distance Comparison [20], and kernel based metric learning [21]. It has been shown that in general, metric learning is capable of boosting the re-identification performance without complicated and handcrafted feature representations. of the current designations.

## III. OUR METHOD

Given a probe image $I_p$ and a set of gallery images $\mathcal{I} = \{I_1, \ldots, I_N\}$, our goal is to determine which element in the gallery set is generated by the same person as $I_p$. To this end, we assign a binary random variable $x_i$, $i = 1, \ldots, N$. for each gallery image to indicate whether it is the target. We also assume that for each gallery image we have extracted a set of feature vectors $\mathbf{f}_i^{(k)}$, $i = 1, \ldots, N$, $k = 1, \ldots, K$, using some feature extraction algorithm. Now we can build a fully connected CRF, as shown in Fig.1. Each node in the CRF corresponds to a label variable $x_i$, and any pair of label variables are connected by an edge to encode the similarity between gallery images. The CRF represents the joint distribution over all labeling variables $\mathbf{X}$ given all observed features $\mathbf{F}$

$$p(\mathbf{X}|\mathbf{F}) \propto \exp\left(-\sum_{i=1}^{N}\phi_u\left(x_i|\mathbf{f}_i\right) - \alpha\sum_{i=1}^{N}\sum_{i<j}\psi_p\left(x_i, x_j|\mathbf{f}_i, \mathbf{f}_j\right)\right) \quad (1)$$

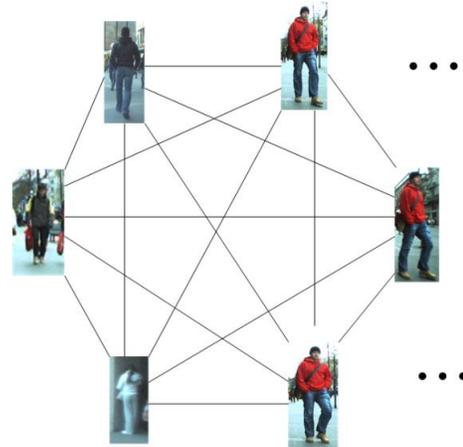

Fig.1. The topology of a full connected CRF. Each image stands for a vertex in CRF and are connected to every other vertices with edges encoding their similarity information.

The unary potential $\phi_u$ encodes the cost of assigning label $x_i$ to node $i$ given observation $\mathbf{f}_i$, and the pairwise potential $\psi_p$ represents the cost of assigning labels $x_i$ and $x_j$ to nodes $i$ and $j$, respectively, given observations $\mathbf{f}_i$ and $\mathbf{f}_j$. Parameter $\alpha$ is a weight learned by cross-validation, controlling the trade-off between unary and pairwise costs.

Now the problem of re-identification transforms to inferring the posterior marginal distribution of each label variable of the CRF. In the next, we will describe what image features we used, how to define the unary and pairwise potentials in CRF, and how to perform efficient inference in the fully connected CRF using technique proposed in [4].

*A. Feature extraction*

Feature extraction aims to encode the image in person-centered bounding box into visual signatures that are more robust and discriminable than raw pixel intensity. Three kinds of features are used in this work for their simplicity and effectiveness, capturing different and complementary characteristic of the image.

**BiCov features** [7]. The first kind of feature used is a combination of biologically inspired features with covariance descriptors. First, biological inspired features are extracted by computing the convolution of original image in 3 color channels (HSV) with a bank of Gabor filters. After applying a max-pooling on the filtered images over two consecutive scales, the resulting magnitude images are encoded by comparing their covariance descriptors at different scales. Finally, the biological inspired features are combined with covariance descriptors to form the image representation. BiCov feature are very robust to illumination variations due to the use of Gabor filter and covariance descriptor. In addition, the max-pooling operation increases the tolerance to scale change and image shift. Furthermore, BiCov can achieve good performance without accurate foreground segmentation as it is based on the difference of covariance descriptor at neighboring scales and the Gabor features and the corresponding covariance descriptors of background regions are very similar.

**wHSV features** [11]. The second feature we used is weighted color histograms which encode the chromatic content of pedestrian in HSV color space. After background subtraction using Stel component analysis [23], the perceptually salient body portions are localized and for each portion a vertical symmetry axis is estimated. Each pixel is weighted according to its distance from the symmetry axis and then the color histogram is constructed for each body portion. The pixels near the symmetry axis are weighted more than that far from it, ensuring to get information from the internal part of the body, trusting less the peripheral portions.

**MSCR features** [22]. The third feature we used is Maximally Stable Color Region (MSCR) features. The MSCR operator detects a set of blob regions that are stable over a range of clustering steps. The descriptor of each region is a 9-dimentional vector containing area, centroid, second moment matrix and average RGB color. MSCR is desirable for person re-identification for its invariance to scale changes and affine transformations of image color intensities.

*B. Potential function*

Now we define the unary and pairwise potentials in CRF.

**Unary potential.** The unary potential measures the cost of assigning a label to a gallery image. We use the distance [7] between probe and gallery features as unary potential. The unary potential of $i$ th node is defined as in Eq.2, where $w_k$ are normalized weights, $\mathbf{f}_p$ and $\mathbf{f}_i$ are features of probe image and the $i$ th gallery image, respectively. $d_{\text{BiCov}}$ is the Euclidean distance in feature space, and $d_{\text{wHSV}}$ is Bhattacharyya distance. Regarding the definition of $d_{\text{MSCR}}$, we use the one given in [11].

**Pairwise potential.** The pairwise potential introduces a penalty when the $i$ th and $j$ th nodes are assigned with different labels. As required by the fast inference algorithm [4], pairwise potential should take the form of a weighted mixture of Gaussian kernels. We define the pairwise potential based on features $\mathbf{f}^{\text{BiCov}}$ and $\mathbf{f}^{\text{wHSV}}$ as in Eq.(3), where $\sigma_m^{(\cdot)}$ are series of kernel width parameters specified experimentally, and $w_m^{(\cdot)}$ are weights of Gaussian kernels that can be learned from training data by solving a least-square problem [CoDet].

$$\phi_u\left(x_i|\mathbf{f}_i\right) \triangleq \begin{cases} w_1 d_{\text{BiCov}}\left(\mathbf{f}_i^{\text{BiCov}},\mathbf{f}_p^{\text{BiCov}}\right) + w_2 d_{\text{wHSV}}\left(\mathbf{f}_i^{\text{wHSV}},\mathbf{f}_p^{\text{wHSV}}\right) + w_3 d_{\text{MSCR}}\left(\mathbf{f}_i^{\text{MSCR}},\mathbf{f}_p^{\text{MSCR}}\right), & \text{if } x_i = 1 \\ 0, & \text{if } x_i = 0 \end{cases} \quad (2)$$

$$\psi_p\left(x_i,x_j|\mathbf{f}_i,\mathbf{f}_j\right) \triangleq \begin{cases} \sum_{m=1}^{M_1} w_m^{\text{BiCov}} k\left(\mathbf{f}_i^{\text{BiCov}},\mathbf{f}_j^{\text{BiCov}};\sigma_m^{\text{BiCov}}\right) + \sum_{m=1}^{M_2} w_m^{\text{wHSV}} k\left(\mathbf{f}_i^{\text{wHSV}},\mathbf{f}_j^{\text{wHSV}};\sigma_m^{\text{wHSV}}\right), & \text{if } x_i \neq x_j \\ 0, & \text{if } x_i = x_j \end{cases} \quad (3)$$

The Gaussian kernels are defined as

$$k\left(\mathbf{f}_i^{(\cdot)}, \mathbf{f}_j^{(\cdot)}; \sigma_m^{(\cdot)}\right) = \exp\left(-\frac{\left\|\mathbf{f}_i^{(\cdot)} - \mathbf{f}_j^{(\cdot)}\right\|^2}{\sigma_m^{(\cdot)}}\right) \quad (4)$$

which can be viewed as a similarity measure between two features. From the definition of pairwise potential we can see that if two gallery images with similar features are assigned with different labels a large penalty will be introduced. Thus the above definition encourages the gallery images labeled as targets to be similar with each others.

### C. Inference algorithm

There are many standard algorithms for solving the inference problem in CRF. However, when the number of nodes is large, as always encountered in practice, the inference of a fully connected CRF is intractable in general. Fortunately, if the pairwise potentials take a form of mixture of Gaussian kernels, the inference can conducted efficiently using the approach described in [4].

The fast inference algorithm [4] is based on a mean field approximation of the CRF distribution, which yields an iterative message passing algorithm for approximate inference. Given current mean field approximation of the marginals $Q_i$, the update equation can be written as

$$Q_i(x_i = l) \propto \exp\left(-\phi_u(x_i = l) - \sum_{l' \neq l} \sum_{j \neq i} Q_j(x_j = l') \psi_p(x_i = l, x_j = l')\right) \quad (5)$$

The key idea underlying [4] is that massage passing in above equation can be viewed as filtering on the field $\{Q_i\}$ with Gaussian kernels in feature space. This enables us to utilize highly efficient approximation for high-dimensional filtering, which reduces the complexity of message passing from quadratic to linear, resulting in an approximate inference algorithm for fully connected CRFs that is linear in the number of nodes and sublinear in the number of edges in the model.

## IV. EXPERIMENTAL RESULTS

### A. Datasets

**ETHZ dataset.** The ETHZ dataset [24], originally proposed for pedestrian detection and later modified for benchmarking person re-identification approaches, consists of three video sequences: SEQ. #1 containing 83 persons (4,857 images), SEQ. #2 containing 35 persons (1,961 images), and SEQ. #3 containing 28 persons (1,762 images). All images have been resized to 64 × 32 pixels. The most challenging aspects of this dataset are illumination changes and occlusions.

**i-LIDS dataset.** The i-LIDS Multiple-Camera Tracking Scenario (MCTS) [25] was captured indoor at a busy airport arrival hall. It contains 119 people with a total of 476 shots captured by multiple non-overlapping cameras with an average of four images for each person. Many of these images undergo large illumination changes and are subject to occlusions. In addition, images have been taken with different qualities (in terms of resolution, zoom level, noise), making very challenging the re-identification over this dataset.

### B. Implementation details

We compute the BiCov feature [7], wHSV feature [11], and MSCR feature [22] using codes provided by the authors. For computational efficiency, we reduce the dimension of BiCov features by PCA before using them. And the weighted HSV features are normalized. We set the weights in Eq.(2) as $w_1 = w_2 = w_3 = 1/3$.

We split dataset randomly into training and test set. For each dataset, we select images of about 1/5 of the total number of persons for the test set while the images of remaining persons are attributed to the training set. The training set $T\{V_i, V_j, GT\}$ consists of image pairs $(V_i, V_j)$ and corresponding labels GT. If $V_i$ and $V_j$ belong to the same person, the ground truth label GT = 1 and 0 otherwise. In practice, the number of negative samples (GT=0) can be much larger than that of positive ones (GT=1). So we down-sample the negative pairs randomly to keep the training set balance. In this paper, the kernel widths in Eq.(3) are set to cover the range of $\lambda * 2^k$ with k varying from –i to j and $\lambda$ is a fine tuning parameter of kernel widths. To learn the weights of Gaussian kernels in Eq.(3), we minimize the distance between the prediction of Eq.(3) and the ground truth GT under the constraint that are weights for convex combination.

In this paper, we pose person re-id as a retrieval problem. Given a probe, we aim to retrieve all the corresponding images from the gallery set. We cannot present the CMC curves like traditional re-id literatures because the main hypothesis underlying CMC is that one and only one item of the gallery could correspond to the probe. Instead, we compute the precision and recall and plot precision-recall curves and use their combination, F score, as the evaluation measures. We compare our method with eBiCov [7] and SDALF [11]. Although the evaluation measure is different, we keep exactly the same parameters as the authors used in their works. We adopt 5-fold cross validation and randomly select instances of each person in test set as probe and report the average performance.

### C. Results and discussion

We present our results in comparison with the baseline method eBiCov[7] and SDALF[11] on datasets in TABLE 1. For each probe we take the max F value and average over all probes and runs. Note that our method outperforms the others on all above datasets. Compared with SDALF and eBiCov, performance improvement of our method can be attributed to the inference in a full connected CRF model, in which the pairwise potential encodes the similarity between each pair of images. Therefore, the accuracy of similarity prediction of Eq.(3) is of vital importance to our method. In Fig.2 we give one instance of the predicted similarity matrix using learned parameters in comparison with ground truth. Note that the predicted similarity depicts a similar pattern to the ground-truth one.

TABLE 1. Comparison of our method with state-of-art methods in averaged F score on different datasets

| Methods | Datasets | | | |
|---|---|---|---|---|
| | ETHZ1 | ETHZ2 | ETHZ3 | i-LIDS |
| SDALF[11] | 76.31 | 74.74 | 85.43 | 48.76 |
| eBiCov[7] | 82.12 | 77.66 | 89.08 | 50.07 |
| Our method | **84.22** | **80.37** | **91.92** | **53.67** |

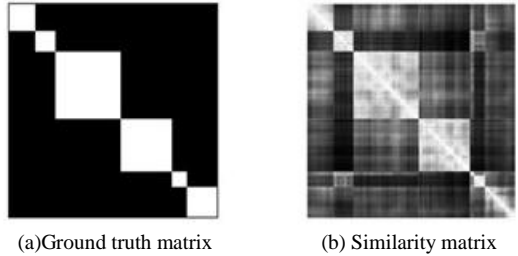

(a)Ground truth matrix     (b) Similarity matrix

Fig.2. Predicting similarity: (a) Ground truth. White stands for positive pairs, black stands for negative ones. (b) Predicted similarity.

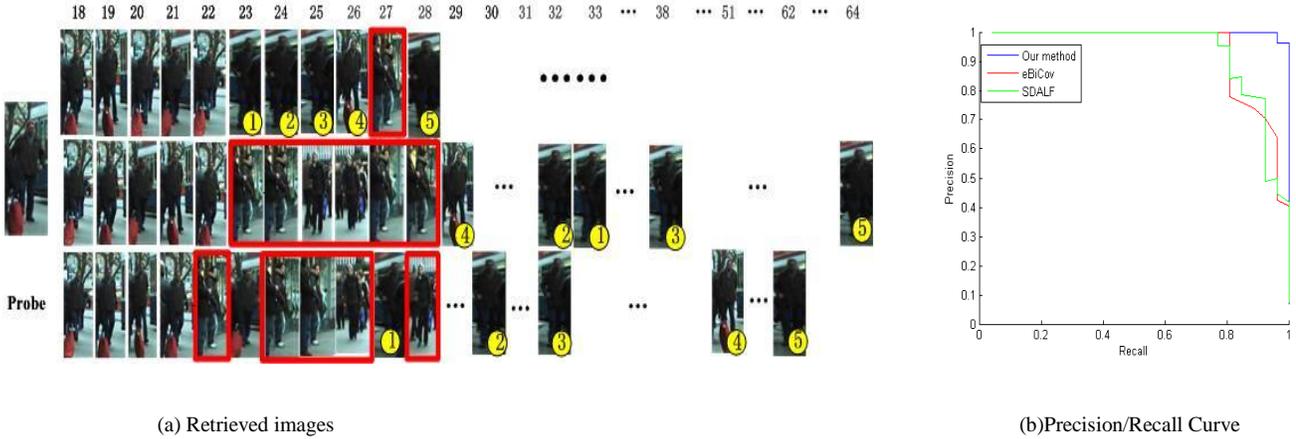

(a) Retrieved images                                           (b)Precision/Recall Curve

Fig.3. An instance on ETHZ dataset. (a)Top row: our method; middle row: eBiCov; bottom row: SDALF. The retrieved images in each row are sorted according to their probability of being target (top row) or their distance from the probe (middle and bottom rows). The mismatched images are marked with red boxes. (b) The corresponding Precision/Recall curve. The images marked with numbered yellow circles are correct matches, the same images in gallery are marked by the same number.

Fig.3 gives a typical instance to illustrate the effect of our method. Fig.3(a) shows the retrieved images using different methods. The images are sorted according to the confidence of being the target. The leftmost is the one with the highest confidence. We only show the images from rank 18 to 64, because the retrieved images by all methods with rank lower than 18 are correct matches. Note that incorrect matches begin to appear from rank 23 in the images sequence retrieved by eBiCov and from rank 22 in the images sequence retrieved by SDALF. We also note that the images marked with yellow numbered circles, which are correct matches of the probe, generally have lower ranks in the images sequence retrieved by our method compared with that by the other two methods. This implies that the inference in full connected CRF, which exploits the similarity information between gallery images, can improve the confidence of true matches which may have large distance from probe in the eBiCov or SDALF feature space.

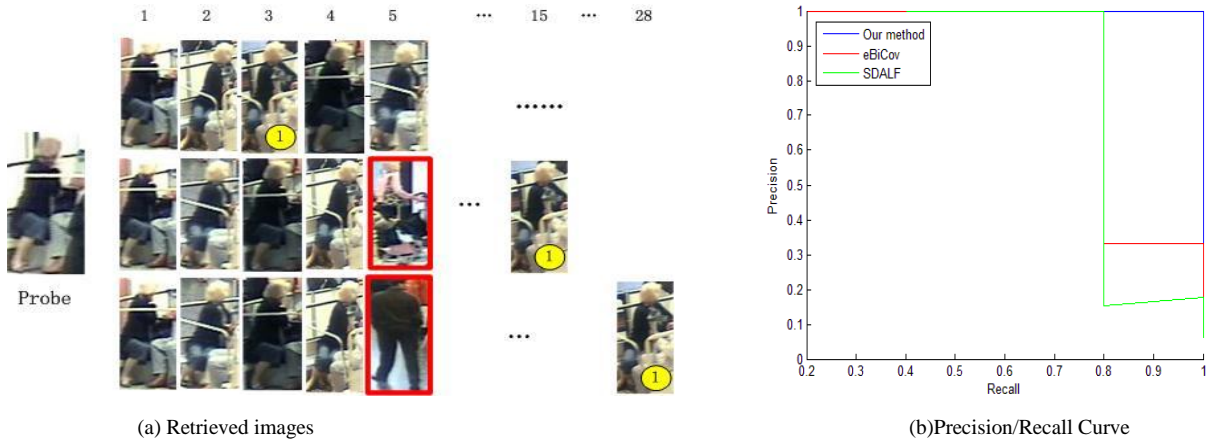

(a) Retrieved images  (b) Precision/Recall Curve

Fig.4. An instance on i-LIDS dataset. (a)Top row: our method; middle row: eBiCov; bottom row: SDALF. The retrieved images in each row are sorted according to their probability of being target (top row) or their distance from the probe (middle and bottom rows). The mismatched images are marked with red boxes. (b) The corresponding Precision/Recall curve. The images marked with numbered yellow circles are correct matches, the same images in gallery are marked by the same number.

We also give a typical instance on i-LIDS dataset in Fig.4. Similarly to ETHZ case, Fig.4(a) shows the retrieved image sequences using different methods sorted according to the confidence of being the target. The leftmost is the one with the highest confidence. There are only 5 images corresponding to the probe. Note that our method can cover all the correct matches within rank 5, while the correct matches marked with yellow numbered circles are retrieved by the other two methods at rank 15 and 28, respectively. This confirms that the inference in full connected CRF can improve the confidence of true matches which may be far from probe in the eBiCov or SDALF feature space by leveraging the similarity information between gallery images.

## V. CONCLUSION

In this paper, we propose a novel method for person re-identification based on inference in fully-connected CRF model. We use recent proposed features [7,11,22] as image signature, and pose the problem of re-identification as a probabilistic inference problem. We exploit the efficient inference algorithm [4] to calculate the marginal distribution of each node. The experimental results show the competitiveness of our method by comparing with the state of the art. Possible extensions include using more effective descriptors, incorporating metric learning [15-21] into the definition of pairwise potentials, and optimizing CRF parameters using max-margin learning [27].

ACKNOWLEDGMENT

This work is supported by the National Natural Science Foundation of China, under grant No.61174020.